\documentclass{article}

    \usepackage[preprint, nonatbib]{neurips_data_2022}

\usepackage[utf8]{inputenc} 
\usepackage[T1]{fontenc}    
\usepackage{hyperref}       
\usepackage{url}            
\usepackage{booktabs}       
\usepackage{amsfonts}       
\usepackage{nicefrac}       
\usepackage{microtype}      
\usepackage{xcolor}         
\usepackage{listings}

\usepackage{floatrow}
\newfloatcommand{capbtabbox}{table}[\captop][\FBwidth]

\usepackage{times}
\usepackage{soul}
\usepackage{url}
\usepackage[tableposition=top]{caption}
\usepackage{float}
\floatstyle{plaintop}
\restylefloat{table}

\usepackage{graphicx}
\usepackage{amsmath}
\usepackage{amsthm}
\usepackage{booktabs}
\usepackage{algorithm}
\usepackage{algorithmic}
\usepackage{subcaption}
\usepackage{enumitem}
\usepackage{wrapfig}
\usepackage{todonotes}
\usepackage{multirow}
\usepackage{colortbl}
\usepackage{hhline}
\usepackage{rotating}

\usepackage{hyperref}       
\hypersetup{
    colorlinks,
    linkcolor={blue!50!black},
    citecolor={blue!50!black},
    urlcolor={blue!80!black}
}

\usepackage{titlesec}
\titlespacing{\paragraph}{%
  0pt}{
  0\baselineskip}{
  1em}

\titlespacing{\section}{%
  0pt}{
  0\baselineskip}{
  0em}

\titlespacing{\subsection}{%
  0pt}{
  0\baselineskip}{
  0em}

\title{SATBench: Benchmarking the speed-accuracy tradeoff in object recognition by humans and dynamic neural networks}

\author{%
 \textbf{Ajay Subramanian} \\
 New York University \\
 \texttt{as15003@nyu.edu}
 \And
 \textbf{Sara Price} \\
 New York University \\
 \texttt{sbp354@nyu.edu}
 \And
 \textbf{Omkar Kumbhar} \\
 New York University \\
 \texttt{omkar.kumbhar@nyu.edu}
 \And
 \textbf{Elena Sizikova} \\
 New York University \\
 \texttt{es5223@nyu.edu} 
\And
 \textbf{Najib J. Majaj} \\
 New York University \\
 \texttt{najib.majaj@nyu.edu} 
 \And
 \textbf{Denis G. Pelli} \\
 New York University \\
 \texttt{denis.pelli@nyu.edu}
}

\begin{document}

\maketitle
\begin{abstract}
The core of everyday tasks like reading and driving is active object recognition. Attempts to model such tasks are currently stymied by the inability to incorporate time. People show a flexible tradeoff between speed and accuracy and this tradeoff is a crucial human skill. Deep neural networks have emerged as promising candidates for predicting peak human object recognition performance and neural activity. However, modeling the temporal dimension i.e., the speed-accuracy tradeoff (SAT), is essential for them to serve as useful computational models for how humans recognize objects. To this end, we here present the first large-scale (148 observers, 4 neural networks, 8 tasks) dataset of the speed-accuracy tradeoff (SAT) in recognizing ImageNet images. In each human trial, a beep, indicating the desired reaction time, sounds at a fixed delay after the image is presented, and observer's response counts only if it occurs near the time of the beep. In a series of blocks, we test many beep latencies, i.e., reaction times. We observe that human accuracy increases with reaction time and proceed to compare its characteristics with the behavior of several dynamic neural networks that are capable of inference-time adaptive computation. Using FLOPs as an analog for reaction time, we compare networks with humans on curve-fit error, category-wise correlation, and curve steepness, and conclude that cascaded dynamic neural networks are a promising model of human reaction time in object recognition tasks. 

\end{abstract}


\section{Introduction} \label{sec:intro}
A fundamental part of everyday tasks like reading and driving is recognizing objects (vehicles, signs, pedestrians while driving; words, letters while reading) and there is often a large variation in the time available to do so. For instance, reaction time permitted while navigating fast-moving, dense traffic is much lower than when driving on empty streets during a global pandemic. Thus, it is important for people to adapt their performance to a wide range of reaction times. When asked to recognize an object, people demonstrate higher accuracy when given more time and can also sacrifice accuracy partially when required to respond quickly. This ability to flexibly tradeoff accuracy for speed is called the speed-accuracy tradeoff (SAT) and is a crucial human skill. 

Recent work has shown that deep convolutional neural networks outperform humans on popular object recognition benchmarks while also being good models for encoding in primate visual cortex \cite{krizhevsky2012imagenet, yamins2014performance}. There is some controversy over the use of neural networks as models of perception. Some models are blueprints that capture all properties of the system. However, modeling in biology is always faced with the complexity of multiple levels of analysis. Biologists are awed by the complexity of neural systems - from synapses to behavior. By necessity, they are reductive scientists, who come up with comprehensible ideas and evaluate how much of the bewildering biological complexity they can model \cite{jonas2017could}. 

Independent of one's position on modeling, it would be valuable for object recognition models to capture how human performance changes as a function of time since it such a salient feature of human behavior. Such models are key to understanding what features are useful for recognition given different reaction times, and how these useful features evolve over time. Towards this objective, we here present a large, public dataset and benchmark of SAT in object recognition by both humans and neural networks. Our human dataset is collected using a reaction time paradigm proposed by McElree \& Carrasco \cite{pmid10641310} where observers are forced to respond at a beep which sounds at a specific time after target presentation. Varying the beep interval across several blocks helps us collect object recognition data across different reaction times.


Standard deep convolutional networks, popular in object recognition, output just a single category prediction for every input. In order to model the temporal dimension, we a) need networks capable of generating outputs at different \textit{timesteps} during inference, and b) have to define \textit{timesteps} for each of them. As candidate models, we consider an emerging class of networks which can vary the amount of computational resources they use during test-time. We chose four of these \textit{dynamic} neural networks that use intermediate classifiers, recurrence or parallel processing as ways to vary computational effort. One of these is a convolutional recurrent neural network (ConvRNN) which has recently been proposed as a model of the human SAT \cite{spoerer2020recurrent}. This network uses lateral recurrence as a way to generate outputs at different \textit{timesteps}. Another network, CascadedNet (referred to hereby as CNet) \cite{iuzzolino2021improving} implements gradual transmission in feedforward networks by taking inspiration from parallel processing in the brain and uses the number of cascaded layers as a measure of time. The other two, MSDNet \cite{huang2017multiscale} and SCAN \cite{zhang2019scan}, are both dynamic-depth networks that were developed towards improving efficiency in computer vision applications. In these, intermediate classifiers in a deep feedforward architecture correspond to different timesteps. We evaluate these networks and humans on the same images and propose three novel metrics to compare them. Our contributions are as follows:
 
 \begin{itemize}[leftmargin=*,itemsep=0pt,topsep=0pt]
  \item We present the first large-scale (148 human observers), public dataset on timed ImageNet \cite{russakovsky2015imagenet} object recognition with 16 categories, across color, grayscale, 3 noise and 3 blur conditions. For each condition, we tested human performance for 5 reaction time (RT) values. This data provides a benchmark for the human SAT and is specifically intended to facilitate comparison between neural networks and humans on timed object recognition.
  \item We present comparable benchmarks for dynamic neural networks, a class of networks capable of inference-time adaptive computation.
  \item We perform an extensive quantitative comparison between SAT in humans and four dynamic neural networks. To do so, we propose three novel metrics: RMSE between SAT curves, category-wise correlation, and steepness which ease model-human comparison. Our dataset\footnote{See \texttt{\url{https://osf.io/2cpmb/}} for dataset.} and code\footnote{\label{sec:website}{See \texttt{\url{https://github.com/ajaysub110/satbench}} for code.}} are publicly available.
\end{itemize}

\section{Related work}

\paragraph{Comparing humans and neural networks.} Human vision inspired early neural networks~\cite{fukushima1988neocognitron,fukushima1982neocognitron} that incorporate some computational features of human vision~\cite{hassabis2017neuroscience}. Many properties of neural networks, such as filters~\cite{bell1997independent} and attention~\cite{lindsay2020attention}, were inspired by the human brain. Recent studies~\cite{zador2019critique} suggest more properties that neural networks might learn from humans, and in this work, we focus on the SAT. We look at the class of networks that can vary their computational effort, and thus model human SAT. In machine learning literature, these models are known as dynamic neural networks~\cite{han2021dynamic}. After a single training procedure, these networks can be evaluated when forced to use different amounts of computational resources (FLOPs)~\cite{hua2018channel,gao2018dynamic,chen2019gater, li2021dynamic,bolukbasi2017adaptive,huang2017multiscale,wang2018skipnet,veit2018convolutional}. Previous work has revealed applications of such networks in resource-sensitive applications such as analysis during autonomous driving~\cite{yang2019re} and mobile health sensors~\cite{xu2019deepwear} are time-sensitive. Being the only class of neural networks capable of inference-time adaptive computation, we attempt to use them to model the SAT in object recognition. We consider a representative sample of dynamic networks which use parallel processing \cite{iuzzolino2021improving}, recurrence \cite{spoerer2020recurrent} or intermediate classifiers \cite{huang2017multiscale, zhang2019scan} as ways to exhibit variable computation.

\paragraph{Measuring the speed-accuracy tradeoff (SAT).} Given more time, people generally do better. McElree and Carrasco~\cite{pmid10641310} analyzed the SAT in humans on a visual search task, in which observers tried to find a target in an array of distractors. They manipulated task difficulty by adding more distractors. Mirzaei et al.~\cite{pmid23419619} proposed a model to predict reaction time in response to natural images. This model is based on statistical properties of natural images and is claimed to accurately predict human reaction time by forming an entropy feature vector. Ratcliff et al.~\cite{pmid14756592} used a drift diffusion model whose drift rate (the rate of accumulation of evidence towards a criterion) was determined by the quality of information to explain lexical decision times and accuracy (i.e. how rapidly does a person classify stimuli as words or non-words). Reaction time has also been studied in the context of perceptual decision making~\cite{palmer2005effect,wong2006recurrent,wagenmakers2007ez,basten2010brain}. Neural networks have been used to model object recognition~\cite{spoerer2017recurrent}, temporal dynamics in the brain~\cite{kietzmann2019recurrence, gucclu2017modeling}, the ventral stream, i.e., the object recognition neural pathway in human cortex~\cite{liao2016bridging}, and temporal information~\cite{bi2020understanding}. Spoerer et al.~\cite{spoerer2020recurrent} used recurrent neural networks to model human reaction times, and were the first to use a neural network as a computational model of the SAT. This work posed a binary classification problem (``animate” vs ``inanimate” objects) to human observers and networks. However, a binary classification task may not represent general categorization accuracy because an observer may learn to detect the difference between categories rather than actually classify images into categories.

\begin{figure}[h]
   \centering
  \vspace{-0.4cm}
  \includegraphics[width=0.7\linewidth]{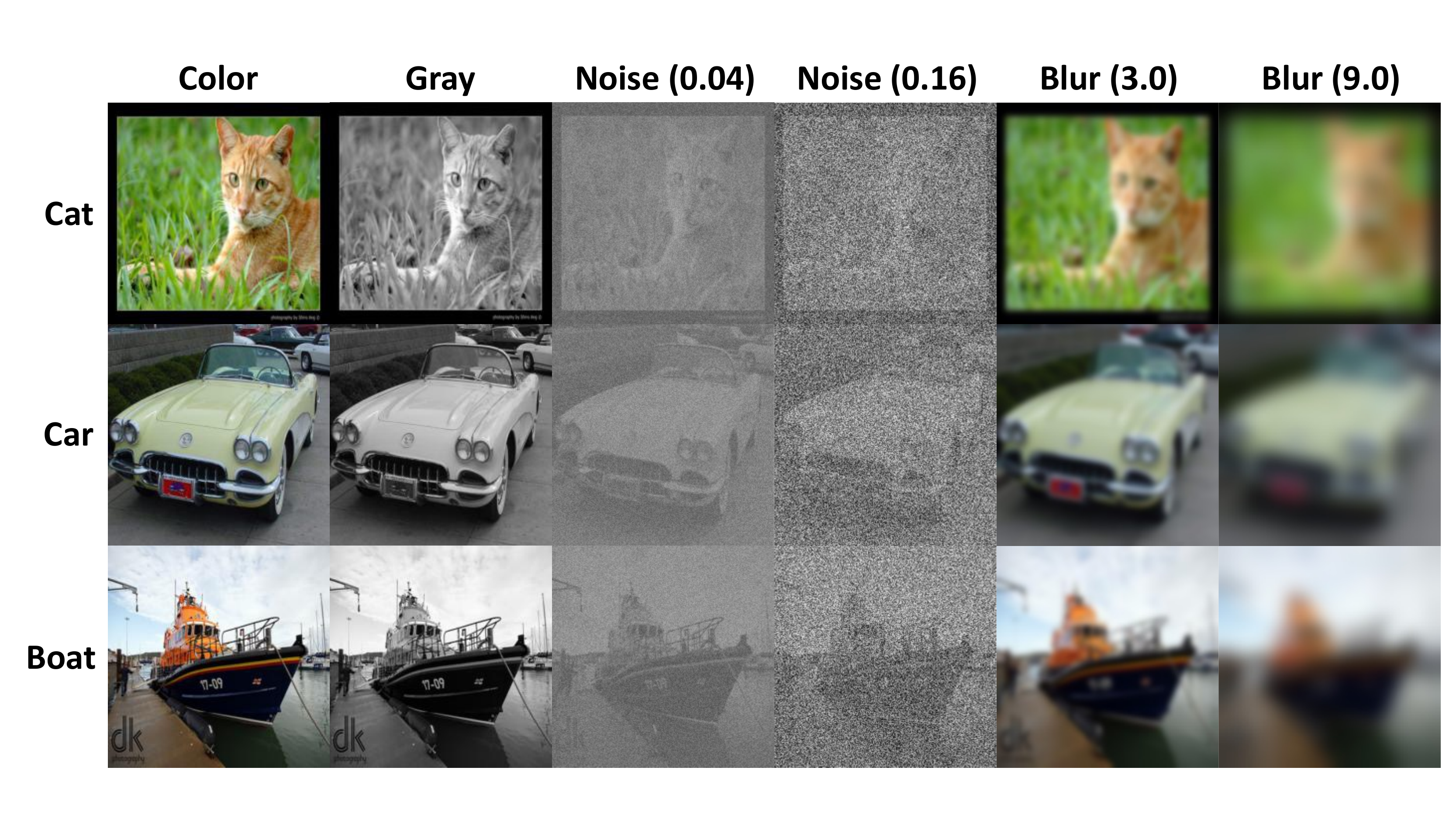}
   \caption{\emph{Sample ImageNet \cite{russakovsky2015imagenet} images shown to observers in our experiment.} Rows correspond to different images, labeled with higher-level category names from 16-class ImageNet \cite{geirhos2018generalisation}. Each column corresponds to a different image transformation used in our experiments. Noise and blur are zero-mean Gaussian and numbers in parentheses indicate standard deviation. Noise was applied to grayscale images after reducing contrast to 20\% (of original) to avoid clipping while blur was applied to the original color images.}
   \label{fig:sample-images}
\end{figure}

\section{Collecting behavioral data} \label{sec:psychophysics}
We measured accuracy and reaction time for human observers performing an object recognition task on images presented with and without perturbation. We assessed the impact of adding color, blur, and noise, The results show a speed-accuracy tradeoff (Figure~\ref{fig:human-network-sat}a.) for all three image manipulations. In Sections \ref{sec:models} and \ref{sec:results}, we evaluate the ability of neural networks to model the tradeoff between processing speed and accuracy. Our experimental protocol is similar to that of McElree \& Carrasco \cite{pmid10641310} and is outlined below. Our dataset contains trial-by-trial information about observer predictions, ground-truth category, image filename, reaction time, etc and therefore can easily be used to study timing-based phenomena in object recognition beyond the SAT.

\paragraph{Images} \label{sec:cifar10} In all our experiments, human observers recognized objects in images from ImageNet \cite{russakovsky2015imagenet}, a popular dataset for neural network analysis. Since memorizing all 1000 categories of the original ImageNet is intractable for human observers, we used 16-class ImageNet \cite{geirhos2018generalisation}, a subset which, using the WordNet hierarchy, is labeled according to 16 higher-level categories: airplane, bear, bicycle, bird, boat, bottle, car, cat, chair, clock, dog, elephant, keyboard, knife, oven and truck. Due to constraints on experiment length, we used a fixed set of 1100 randomly sampled images for all experiments. Figure \ref{fig:sample-images} shows sample images under all viewing conditions used in our experiments. Images were resized from 224x224 to 400x400 pixels for optimal viewing \cite{Pelli_1999}. We estimate the size in cm of the 400x400 pixel image to be 4x4 cm, subtending 4x4 deg, and the viewing distance (distance between observer eye and screen) to be roughly 57 cm. Since additive noise is ill-defined for color images, we converted images to grayscale for all noise experiments after also reducing contrast to 20\% of original to avoid clipping at floor, ceiling. For analysis verifying that both of these modifications do not significantly degrade human performance, see Supplementary Material.

\paragraph{Observer statistics and data collection.} \label{sec:user_study}

\begin{wraptable}{r}{0.4\textwidth}
\vspace{-0.3cm}
\begin{center}
\resizebox{1.0\textwidth}{!}{
\begin{tabular}{lllll} 
\toprule
\multirow{2}{*}{Exp.} & \multirow{2}{*}{\#Partic.} & \multicolumn{2}{l}{Compl. time (min.)} & \multirow{2}{*}{\#Trials}  \\                                                                      
                            &                                 & Mean    & SD                               &                            \\ 
\cline{1-5}\\
Col. \& Gr.                      & 58                              & 43.24~  & 13.32                             & 1100                        \\
Gray Noise                       & 45                              & 41.32~  & 11.25                            & 1100                       \\
Color Blur                        & 45                              & 39.29 ~ & 6.66                            & 1100                       \\
\bottomrule
\end{tabular}
  }
\caption{\emph{Summary statistics of collected data on human observers across all experiments.}}
\label{tab:summary-stats}
\end{center}
\end{wraptable}

We collected data from a total of 148 participants, recruited through Amazon MTurk \cite{crowston2012amazon}. Each session (set of trials) lasted about an hour. Each observer had a normal or corrected-to-normal vision. The stimuli were presented via a JATOS survey via worker links to each observer. Participants were paid \$20 for their efforts and the total cost of data collection was \$3860. A standard IRB-approved (IRB-FY2016-404) consent form was signed before collecting the data by each observer, and demographic information was collected. Please refer to Supplementary Material for a detailed description and documentation of our dataset.

\paragraph{Survey design.} The survey was designed to control the response time of human observers by asking them to respond in the allotted time distribution. The design was based on previous work by McElree \& Carrasco \cite{pmid10641310}. In our case, 3 separate experiments were run for color/gray, noise and blur. Each experiment consisted of an initial training block of 50 trials followed by 5 blocks of 210 trials, each corresponding to a different reaction time. The training trials were used to train observers to click one of 16 virtual buttons (corresponding to the 16 image categories) in response to the stimulus image. During the reaction time blocks, each trial involved an image being presented and required the observer to click their response at a timed beep. Each of the 5 timed blocks used a different interval between the stimulus and beep: 500, 900, 1100, 1300, and 1500 ms. An additional 200 ms was given after the beep for slightly delayed responses. All observers saw the same set of images within each reaction time block, in random order. Number of images from all categories was approximately equal for any given block. To avoid order effects in our results, an observer was equally likely to view the timed blocks either in ascending or descending order of reaction time. The first 10 trials of each block were discarded before analysis because they were assumed to be used by observers to adapt to the block's beep timing. Observer data with more than 50\% of data outside a $\pm100$ ms range of the beep were discarded.

\emph{Note.} The McElree \& Carrasco SAT paradigm~\cite{pmid10641310} was a major advance in tracking the improvement of accuracy with time. An alternative approach of allowing observers to respond when they feel like and then sorting into bins produces confounds that make the data hard to analyze because observers tend to take longer on harder trials. In our case, we trained observers to respond at a fixed time (different in each block), so measured accuracy is not confounded with trial-by-trial difficulty. Our use of their paradigm makes our results much easier to analyze.  In many studies of the effect of timing in object recognition~\cite{majaj2015simple,rajalingham2018large,tang2018recurrent,geirhos2018generalisation}, each trial’s stimulus presentation and choice selection are separate steps. Various stimulus durations are reported: 100-2000 ms in \cite{majaj2015simple}, 100 ms in \cite{rajalingham2018large}, 25-150 ms in \cite{tang2018recurrent}, 200 ms in \cite{geirhos2018generalisation}, after which the observers are allowed to take as much time as needed to make their selection. In our experiments (the SAT paradigm), each trial was one step. The image stayed on until the observer responded. Thus, our reported reaction times include all the time between stimulus onset and key press. Our observers had very little time to respond, compared to typical object recognition studies, and as a result, their accuracy appears lower than of those from other studies. In our case, the lowest timing threshold was specifically restricted so that the human accuracy is near chance. 

During the experimental session, observers were instructed to try their best to respond at the beep, were given feedback after every trial and were continuously presented with a trial progress counter. Clicking on the fixation cross using their mouse would invoke the next trial. This ensured that the observer fixated at the center before a trial and also that they would have to move the mouse pointer from the same location before response, thus maintaining almost constant motor-time.

\begin{figure}[h]
   \centering
  \vspace{-0.37cm}
  \includegraphics[width=0.75\linewidth]{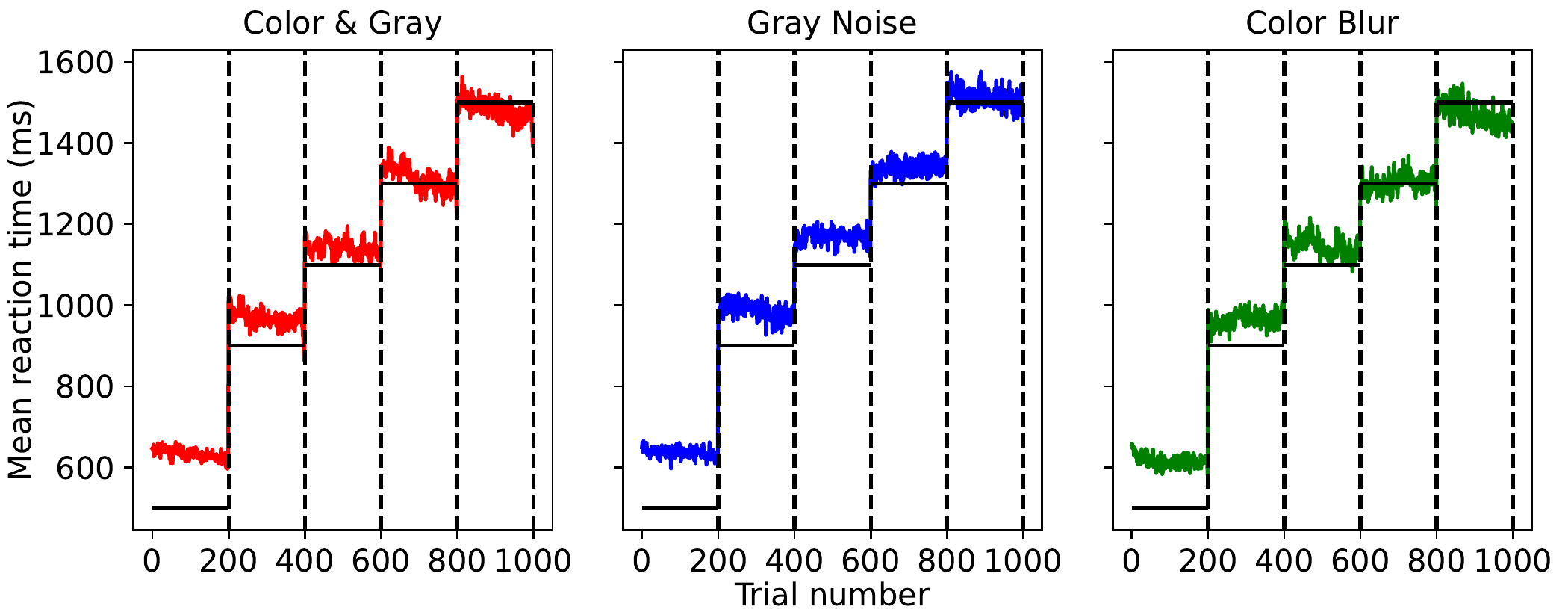}
   \caption{\emph{Reaction time (RT) for each timed trial in each experiment, averaged across human observers.} Vertical dotted lines separate different RT blocks, each of which had 200 trials after discarding the first 10 used for training. Horizontal black lines indicate beep timing at which observer was asked to respond and colored lines denote observer's reaction time. Approx. 50\% of observers saw trials in descending order of RT blocks, but we reverse their trial numbers here for the purpose of averaging and visualization.}
   \label{fig:human-rt}
\end{figure}

\begin{figure}
   \centering
  \vspace{-0.4cm}
  \includegraphics[width=0.8\linewidth]{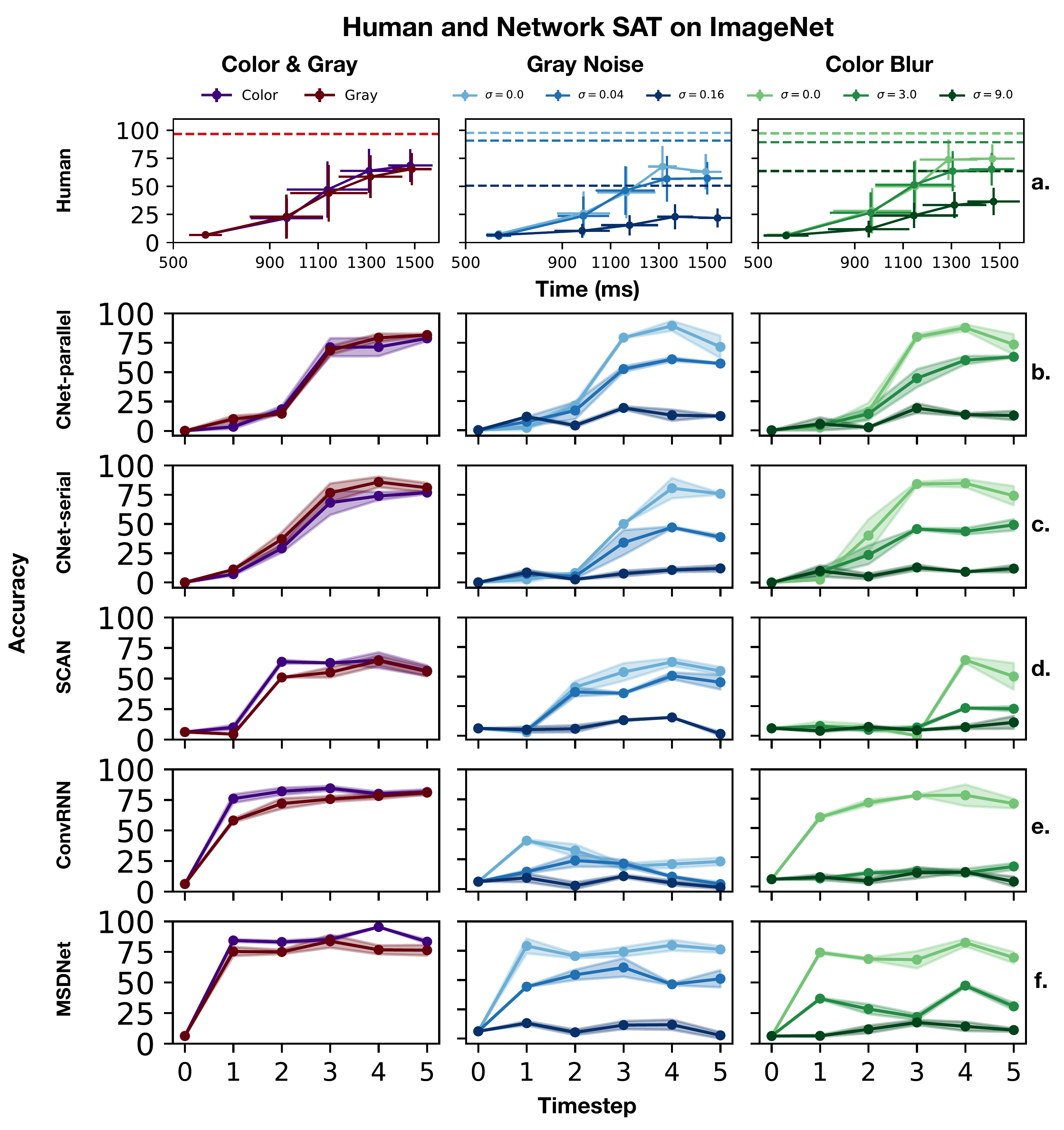}
   \caption{\emph{Accuracy vs reaction time curves for humans and dynamic neural networks.} Each point corresponds to average accuracy and reaction time for a single RT block, with error bars showing standard error across both axes. Each plot shows curves for all image transforms in the experiment in different colors. \textbf{a.} Dotted horizontal lines of same color denote accuracy for untimed condition, averaged across training trials. Humans show a large range of accuracies over RTs and a gradual trend, two desirable aspects that we seek to model. \textbf{b-f.} For networks, the "Time (ms)" axis is replaced with "Timestep" which is defined independently for each network (see Section \ref{sec:models}). Accuracy at timestep 0 is assumed to be at chance.}
   \label{fig:human-network-sat}
\end{figure}

\paragraph{Data quality} Given that the observers had to identify complicated images perturbed with noise/blur while trying to respond exactly at the beep, the task was particularly difficult which made it necessary to verify data quality. Figure \ref{fig:human-rt} shows the distribution of reaction time across trials in all 3 experiments, averaged across participants. On average, our participants responded very close to the beep across all blocks and displayed high consistency across trials within each block. We also observed little difference between data collected online and very similar data from experiments with in-person testing when using CIFAR-10 images ~\cite{kumbhar2020anytime, Krizhevsky09learningmultiple}.

\section{Modeling the speed-accuracy tradeoff (SAT) with dynamic neural networks} \label{sec:models}

Having collected a dataset that demonstrates a strong SAT in humans (Figure \ref{fig:human-network-sat}a.), we now move to using neural networks to capture this flexible, adaptive behavior. Dynamic neural networks are a class of networks that are capable of inference-time adaptive computation. Test performance can be obtained at different amounts of computation after a common training procedure. This is similar to the paradigm we used to evaluate human observers, with computational resources in networks used as an analogue for human reaction time. Only a limited amount of existing work develops such dynamic architectures, in contrast to the abundance of networks available for non-temporal object recognition \cite{geirhos2021partial}. We consider 4 network architectures that represent the variety of adaptive computation approaches and give a brief overview below. For detailed description of architecture and training procedure, please refer to the Supplementary Material.

\begin{itemize}[leftmargin=*,itemsep=0pt,topsep=0pt]
\item\textbf{Convolutional Recurrent Neural Network (ConvRNN)} ~\cite{spoerer2020recurrent} exhibits temporal behavior by relying on lateral recurrent connectivity, characteristic of the primate visual system, implemented by adding layer-wise feedback connections to a feed-forward convolutional network. This model consists of several blocks of recurrent convolutional layers, followed by a readout layer to output category predictions. During inference for a given input image, the computation used by the model can be dynamically selected by running the network for a variable number of recurrent cycles. This property allows the network to respond to an input image with different numbers of forward passes, which we use as an analogue for reaction time.

\item\textbf{Multi-Scale Dense Network (MSDNet)} ~\cite{huang2017multiscale} implements dynamic inference using multiple (5, in our case) intermediate classifiers from a feedforward network. Since these early-exits are all at different depths, classification at each one has a different computational requirement. Since all exits use features from a common backbone network, there is an interference between the features deemed useful for classification at each depth. To resolve this problem, MSDNet proposes two architectural features: multi-scale feature maps, and dense connectivity (realized by using a DenseNet~\cite{Huang_2017_CVPR} backbone). These properties allow neurons at any layer to access features from any part of the network and at any resolution, thus diminishing the effect of the interference problem. We refer to the index of an exit as the timestep corresponding to its prediction because the 5 exits are approximately evenly spaced along the backbone.

\item\textbf{Scalable Neural Network (SCAN)} ~\cite{zhang2019scan}, like MSDNet, implements dynamic inference using early exit classifiers from a common backbone network. Whereas MSDNet uses multi-scale feature maps and dense connectivity to solve the issue of interference between early and late classifiers, SCAN uses an encoder-decoder attention mechanism in each exit network. This allows each exit to ``focus'' only on features relevant for classification at a specific depth. The attention network produces a binary mask which is added to the backbone (ResNet~\cite{he2016deep}) feature map, after which a Softmax layer predicts a class label. The network uses four early exits and a final ensemble output which uses all early exit features for prediction. Thus, for a given input, the network outputs five class predictions, each requiring a different amount of computation time/effort.

\item\textbf{Cascaded Neural Network (CNet)} \cite{iuzzolino2021improving} uses skip connections and parallel processing in a ResNet \cite{he2016deep} backbone to implement gradual transmission of activity. The output at the $i$th timestep uses activation at the $(i-1)^{th}$ ResNet block as well as partial activations from previous timesteps to simulate a cascading effect. Additionally, it utilizes a temporal difference (TD) loss whereby the target for the prediction at each timestep is the discounted sum of targets at future timesteps. The ground truth label is used as the target for the final timestep. The version of CNet that implements both TD loss and parallel processing is henceforth referred to as `CNet-parallel'. We also consider an ablation called `CNet-serial' that also makes use of the TD loss but with no parallel processing i.e., the output at timestep $i$ is just the output when $i-1$ blocks of the ResNet compute an activation, which goes straight to the final fully-connected layer.
\end{itemize}

All networks were trained from scratch on the ImageNet dataset to perform the aforementioned 16-way categorization task, following which they were all evaluated on the same images used in our human experiments. Since the total number of timesteps for all networks, 5, is equal to the number of reaction time values we considered, we map them linearly and use the test images for each RT block to evaluate the networks at the corresponding timestep. We also include an additional sixth data point for all models, assuming chance accuracy ($6.25$\%) at timestep 0 i.e., when no computational resources are used.

\section{Results} \label{sec:results}
Figure \ref{fig:human-network-sat} shows the accuracy vs. time/timestep curves for humans and networks. Human accuracy gradually grows as RT is increased, while accounting for a large range of accuracy values. Peak accuracy drops when noise/blur is added, while maintaining an increasing trend with RT. Also, humans perform significantly better in the untimed condition (accuracy over training trials). We hypothesize that this effect is due to not only more time, but also due to less cognitive load when observers are asked to respond without consideration to beep timing. Individual SAT curves for each human observer are shown in Supplementary Material. CNet-parallel and CNet-serial qualitatively capture the trends of the human curves. They gradually increase over a large range of accuracies. SCAN's curves increase slowly but over a much smaller range. MSDNet's trend is very steep and shows almost no change in accuracy with time. In service of a more objective analysis, we compared humans and networks using 3 metrics: curve-fit error, category-wise correlation, and curve steepness.

\begin{figure}[h]
   \centering
  \vspace{-0.4cm}
  \includegraphics[width=0.75\linewidth]{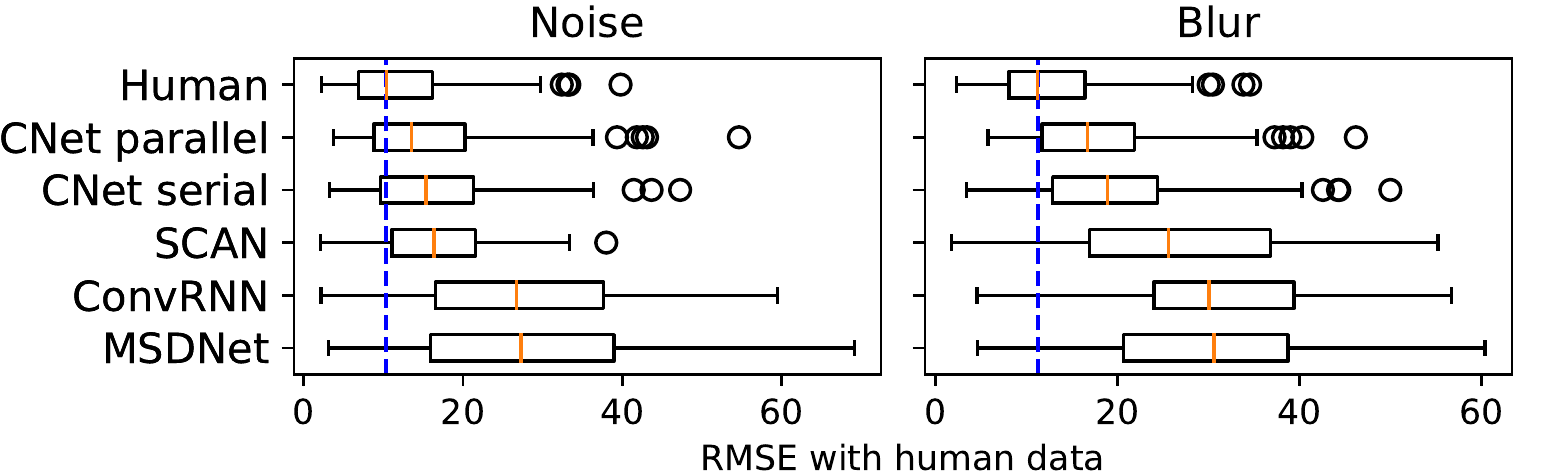}
   \caption{\emph{Boxplot showing RMSE of curve-fits of model data to human SAT.} ``Human'' represents RMSE when the average human curve is fit to each human observer. Orange marker is median RMSE. Blur dotted line is an extension of the human median.}
   \label{fig:fig5}
\end{figure}

\textbf{Curve-fit error: How well do network curves match human data?}
To test how well network curves match the human SAT, we computed the root-mean-squared-error (RMSE) between network curves and the SAT curve of each human observer. Since the human and network (without the assumed timestep 0 point) curves have equal number of points, this was easily calculated. RMSE was found separately for each noise/blur value and then averaged. We did so because we wanted a single metric to represent how well each model captures performance across \textit{all} values of noise or blur. 
\begin{align*}
    e_{RMSE}(c_1,c_2) = \frac{1}{N_p} \sum_{p=1}^{N_p} \sqrt{\frac{1}{N_t} \sum_{t=1}^{N_t} (c_1[p,t] - c_2[p,t])^2}
\end{align*}
where $e_{RMSE}$, the curve-fit error for two curves $c_1, c_2$, each defined over $N_p$ noise/blur perturbations and $N_t$ reaction times, is the RMSE computed across time and averaged across perturbation curves.
Figure \ref{fig:fig5} summarizes our results in a boxplot across both noise and blur experiments. We also include ``Human'' which denotes the RMSE when the average human SAT curve was fit to each individual observer. We observe that across noise and blur experiments, both CNet-parallel and CNet-serial achieve very low RMSE with human data while also capturing the distribution well. SCAN's error is low only in noise and much higher for blur where it is similar to MSDNet and ConvRNN which provide the least accurate fits across both experiments.

\textbf{Category-wise correlation: How similarly do humans and networks behave across object categories?} Having shown that cascaded networks, to a large extent, capture the SAT behavior in humans, we move to analyzing how the behavior of networks matches with humans across categories. Do the network and human curves resemble each other for the same categories? To understand this, we measured the category-wise correlation of each network to human data. To do so, we obtained separate SAT curves for each category, flattened them into a single vector and then computed Spearman's rank correlation between vectors for networks and each human observer.

\begin{figure}[h]
   \centering
  \vspace{-0.4cm}
  \includegraphics[width=0.6\linewidth]{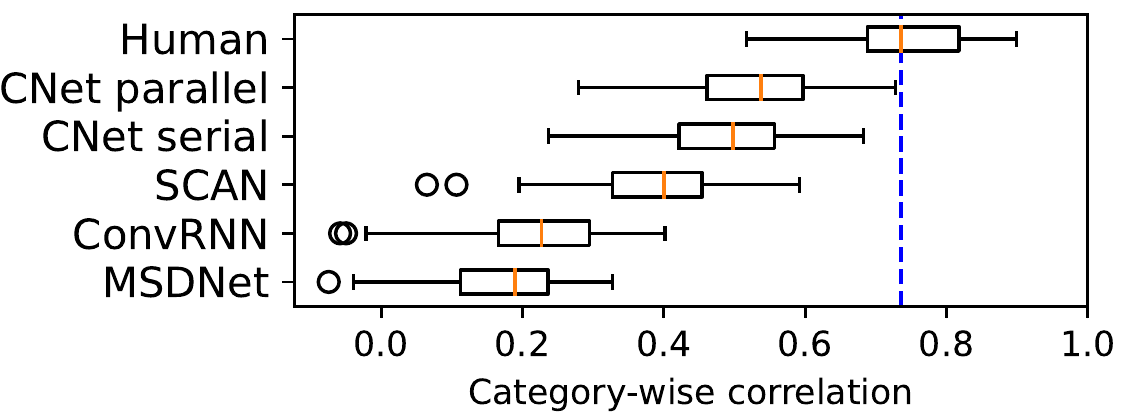}
   \caption{\emph{Boxplot showing category-wise Spearman's rank correlation between curve-fits of model data and human SAT in the color experiment.} ``Human'' represents category-wise correlation between the average human curve and each human observer. Orange marker is median correlation. Blue dotted line is an extension of the human median. Category-wise correlation is found by computing correlation between 1-D vectors containing category-wise SAT curves flattened across categories.}
   \label{fig:category-correlation}
\end{figure}

\begin{figure}[h]
   \centering
  \vspace{-0.4cm}
  \includegraphics[width=\linewidth]{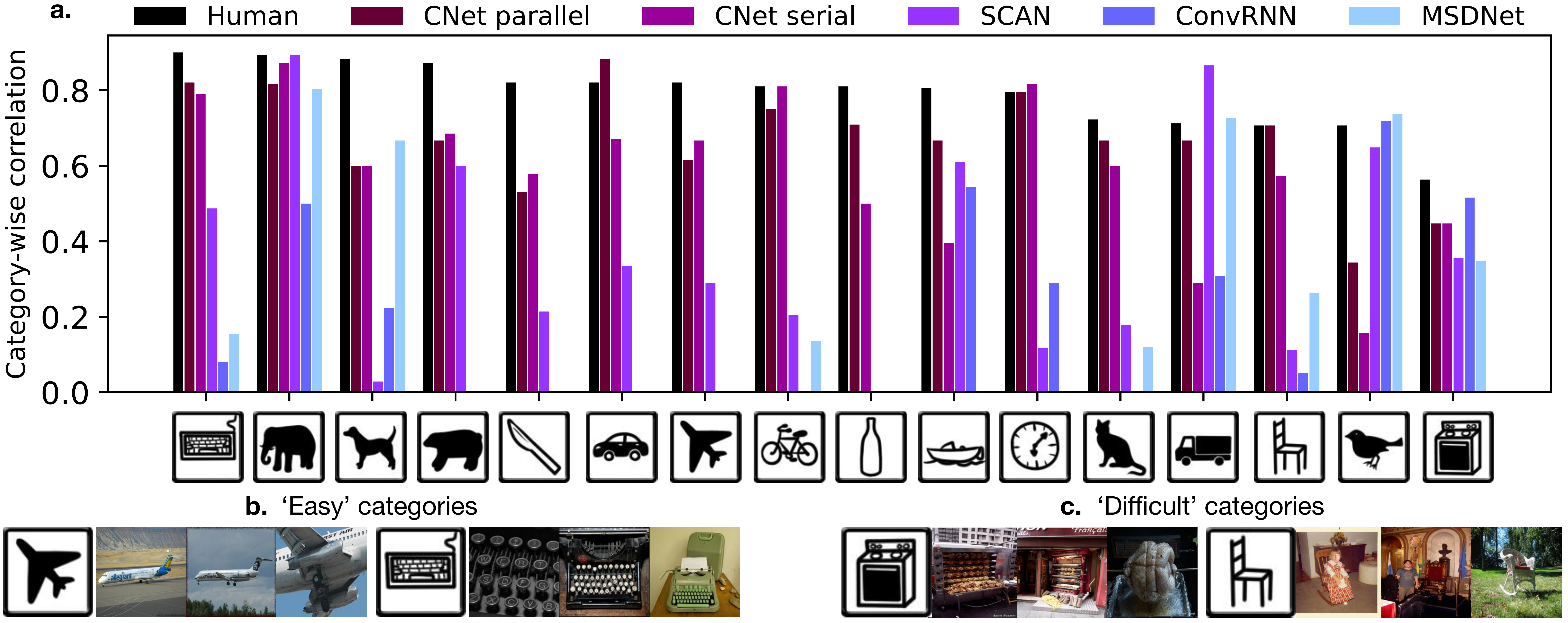}
   \caption{\emph{Analysis of network-human correlation across individual categories.} \textbf{a.} Barplot showing network-human correlations separately for each category. Y-axis represents median (across human observers) correlation between network and human SAT curves for a given category. Correlations were computed for all models and categories but negative values were clipped for the sake of visualization. Icons on the x-axis are illustrations of the 16 categories, borrowed from \texttt{https://cocodataset.org/\#explore}. \textbf{b.,c.} Sample images from categories for which the best model, CNet-parallel, obtained high and low correlations are shown as `Easy' and `Difficult' categories respectively.}
   \label{fig:category-correlation-bar}
\end{figure}

Figure \ref{fig:category-correlation} illustrates our results as a boxplot. The CNets are the only models achieving >0.5 correlation. However, the gap between human and CNet performance has increased relative to Figure \ref{fig:fig5}. To determine a possible cause for this change, we look at Figure \ref{fig:category-correlation-bar}a. which shows a separate median correlation barplot for each category. We see that the CNets show a positive correlation larger than 0.5 for most categories while MSDNet, ConvRNN and SCAN correlate poorly with several categories and sometimes also show negative correlations (clipped in figure for purpose of visualization). For the best model, CNet-parallel, 'oven' and 'chair' are the only cases where correlation is below 0.5. We suspect that this is due to the fact that images with chairs in ImageNet often contain other objects which could confuse human observers, and the oven class is ambiguous given that microwave ovens, grills and woodfire ovens, etc. are all labeled ovens, which makes it a more difficult category for humans than CNets (Figure \ref{fig:category-correlation-bar}c.). `Easy' categories like airplane and keyboard, on the other hand, are less ambiguous and hence, both humans and CNet-parallel perform similarly (Figure \ref{fig:category-correlation-bar}b.). `Easy' classes are also generally highly correlated to the average human, while `Difficult' classes show low correlation, signaling higher ambiguity. Both kinds of categories hardly affect peak network accuracy and hence, the mismatch with humans. Figures \ref{fig:category-correlation-bar}b. and \ref{fig:category-correlation-bar}c. illustrate this hypothesis using randomly sampled images from categories for which CNet-parallel showed high and low correlation.

\textbf{Curve steepness: How sharply does accuracy change as a function of reaction time?} \label{sec:steepness} A characteristic feature of the human SAT is that it is gradual \cite{fitts1966cognitive, wickelgren1977speed}. People fail gracefully as reaction time is decreased. We use mean curvature of the cumulative-Weibull fit as a measure of curve steepness and compare humans with networks. We first fit the cumulative Weibull function to network and human curves, since previous work has used it to describe the SAT \cite{van1991rasch}. Then, we measure mean curvature which computes on average how drastically the slope of the tangent to the curve changes between successive points. We refer to this metric as steepness. For a detailed explanation of how steepness is computed, please refer to the Supplementary Material.

\begin{figure}[h]
   \centering
  \vspace{-0.4cm}
  \includegraphics[width=0.65\linewidth]{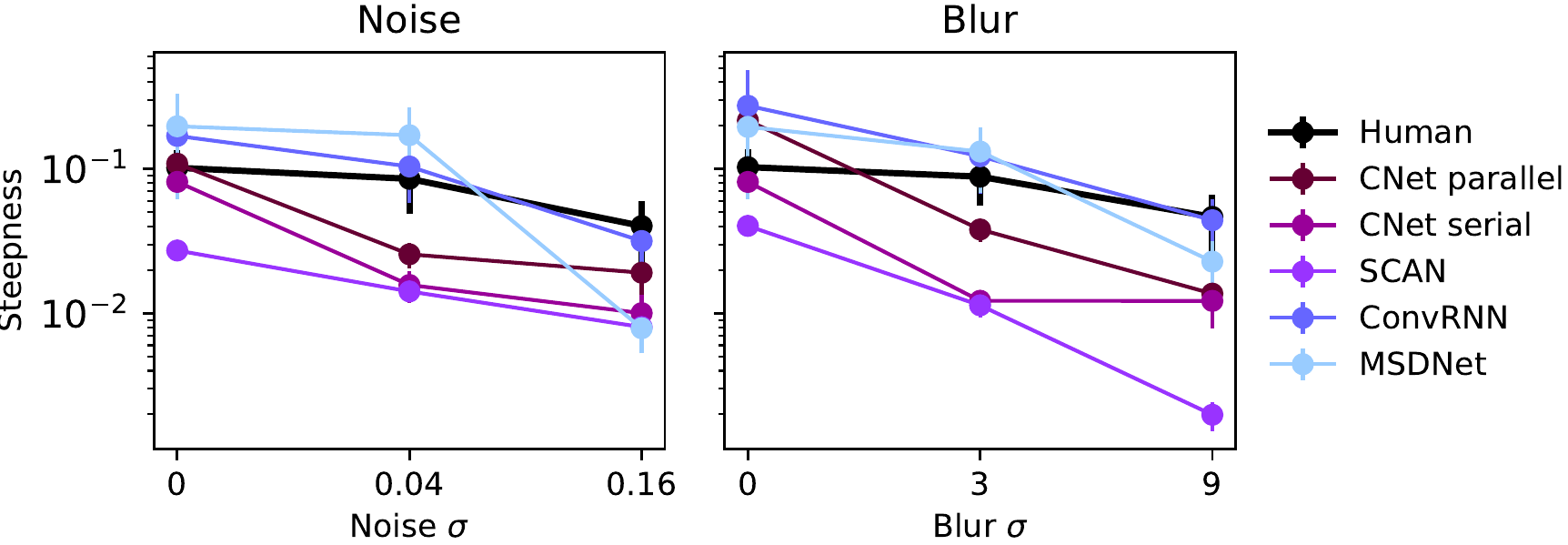}
   \caption{\emph{Steepness of human and network SAT curves across noise and blur conditions, shown in a log-scale} Humans show low steepness that stays almost flat across all conditions. CNets resemble human steepness range. MSDNet and SCAN show much higher and much lower steepness than humans respectively. For mathematical definition of steepness, refer to Section \ref{sec:steepness}.}
   \label{fig:fig7}
\end{figure}

Figure \ref{fig:fig7} plots steepness across different noise/blur values for models and average human, on a log scale. Our main observation is that human steepness is almost constant across all noise and blur conditions, similar to previous observations \cite{nachmias1981psychometric}. This shows that people's failure rate when time is decreased is independent of task difficulty. MSDNet is, on-average, steeper, changing very little between low noise/blur values but significantly between mid-high values. This is expected since the MSDNet curves in Figure \ref{fig:human-network-sat}e show a very steep rise for low noise/blur conditions. SCAN shows significantly lower steepness than humans in both noise and blur. This is due to the fact that SCAN covers a small range of accuracies by increasing very gradually. In general, none of the networks adequately capture the flatness of the human psychometric curve.

\section{Conclusion} \label{sec:conclusion}
The SAT is an important feature of human performance that is difficult to explain with current computational models of object recognition. We have collected a large dataset of SAT by humans recognizing ImageNet images, across color, grayscale, noise and blur conditions. In these data, people are more accurate when given more time, across all conditions, demonstrating a large range of accuracies (chance - 75\%) and a graceful trend. Both range and steepness were therefore properties that we sought to model using neural networks. In order to compare network and human performance, we proposed 3 metrics: curve-fit error, category-wise correlation and curve steepness which allow for a robust analysis. We determined that cascaded dynamic neural networks (CNets) can, to a high degree, capture the human SAT across all noise and blur conditions while also displaying high category-wise correlation. However, none of the networks can adequately capture the gracefulness of the human curve, as measured by our steepness metric. Humans fail at the same rate across all noise and blur conditions while network curve steepness changes significantly. This leaves plenty of space for future work on our benchmark.

What properties of networks could allow for a human-like SAT and what are the specific correlates, if any, between these architectures and the brain? Also, how do the features and representations used by the brain change with time? We hope that our dataset and comprehensive benchmark will encourage future work that explores these crucial questions towards bridging the gap between human and machine vision. Applications of the above-described technology have potential benefits (addressing public health concerns — e.g., slow reading — and biases in computational models) and risks (facilitating the creation of bots that pass for humans for malicious purposes). Such concerns are shared by much research in computational modeling, and are outside the scope of this work.

\section{Acknowledgements}
The authors would like to thank Jean Ponce, Michael Picheny, Jack Epstein, Kieran Sim, Pedro Galarza, and anonymous reviewers. We acknowledge support from NIH grant 1R01EY027964 to Denis Pelli and the Moore-Sloan Data Science Environment initiative (funded by the
Alfred P. Sloan Foundation and the Gordon and Betty
Moore Foundation through the NYU Center for Data Science) to Elena Sizikova. We thank staff at the high-performance computing facilities at NYU for providing support and compute resources.

\newpage
{\small
\bibliographystyle{unsrt}
\bibliography{bibliography}
}

\appendix

\newpage
\centerline{\Large{\textbf{Supplementary Material}}}

\section{Behavioral Data Collection}
\begin{figure}[h]
   \centering
  \vspace{-0.4cm}
  \includegraphics[width=\linewidth]{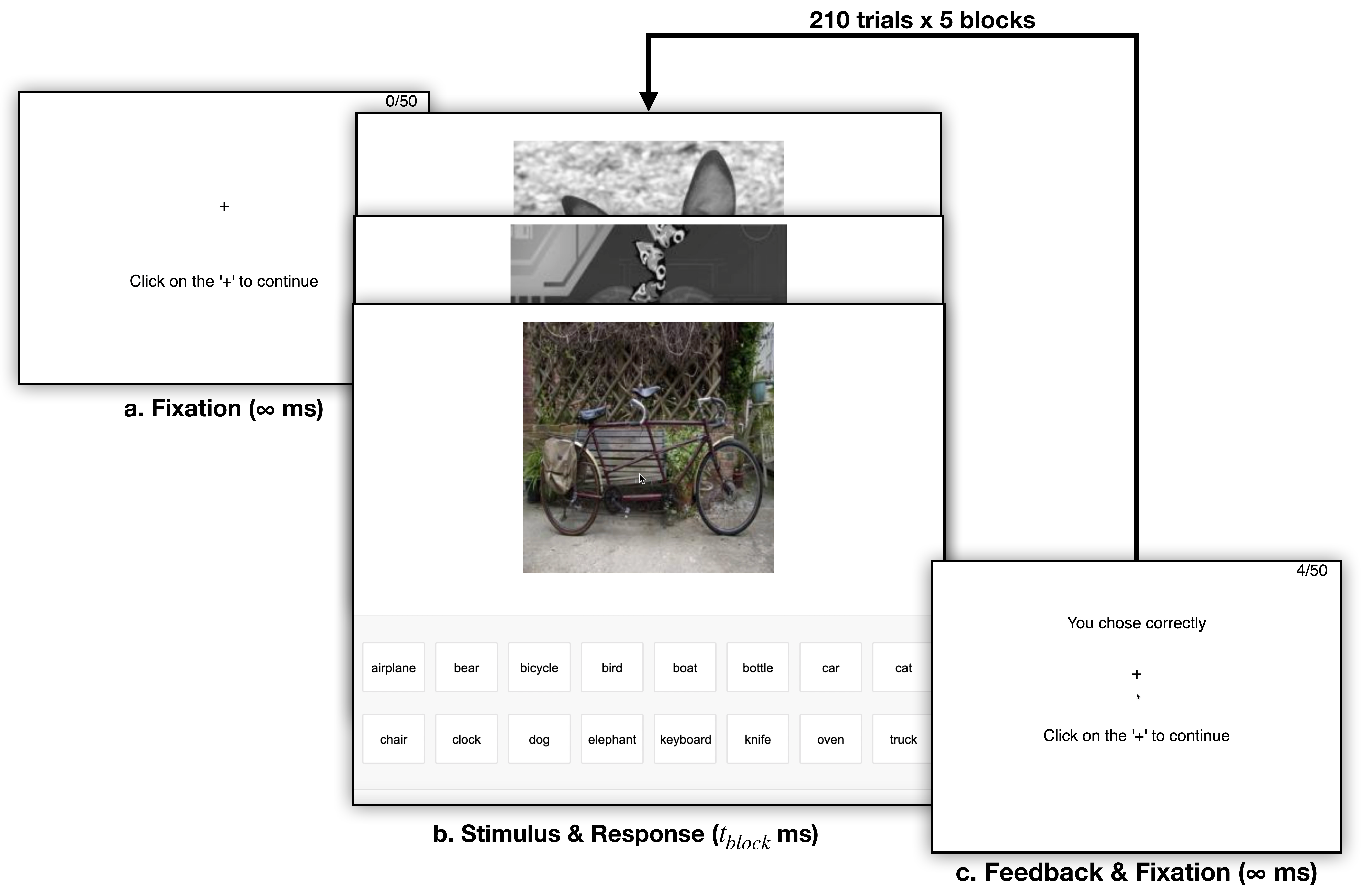}
   \caption{\emph{Sequence of screens in each trial of our psychophysical experiment.} \textbf{a.} Each block in our experiments begins with a fixation screen which prompts the observer to click on a central cross. This is done to ensure that the observer fixates at the center of the image at the beginning of each trial. Clicking the cross presents the \textbf{b.} stimulus \& response screen. An image and 16 category buttons below it are displayed. The observer is asked to select a category when a beep sounds at $t_{block}$ ms. \textbf{c.} Then, the next screen gives feedback and solicits fixation for the next trial.}
   \label{fig:screenshots}
\end{figure}

\begin{figure}[h]
   \centering
  \vspace{-0.4cm}
  \includegraphics[width=\linewidth]{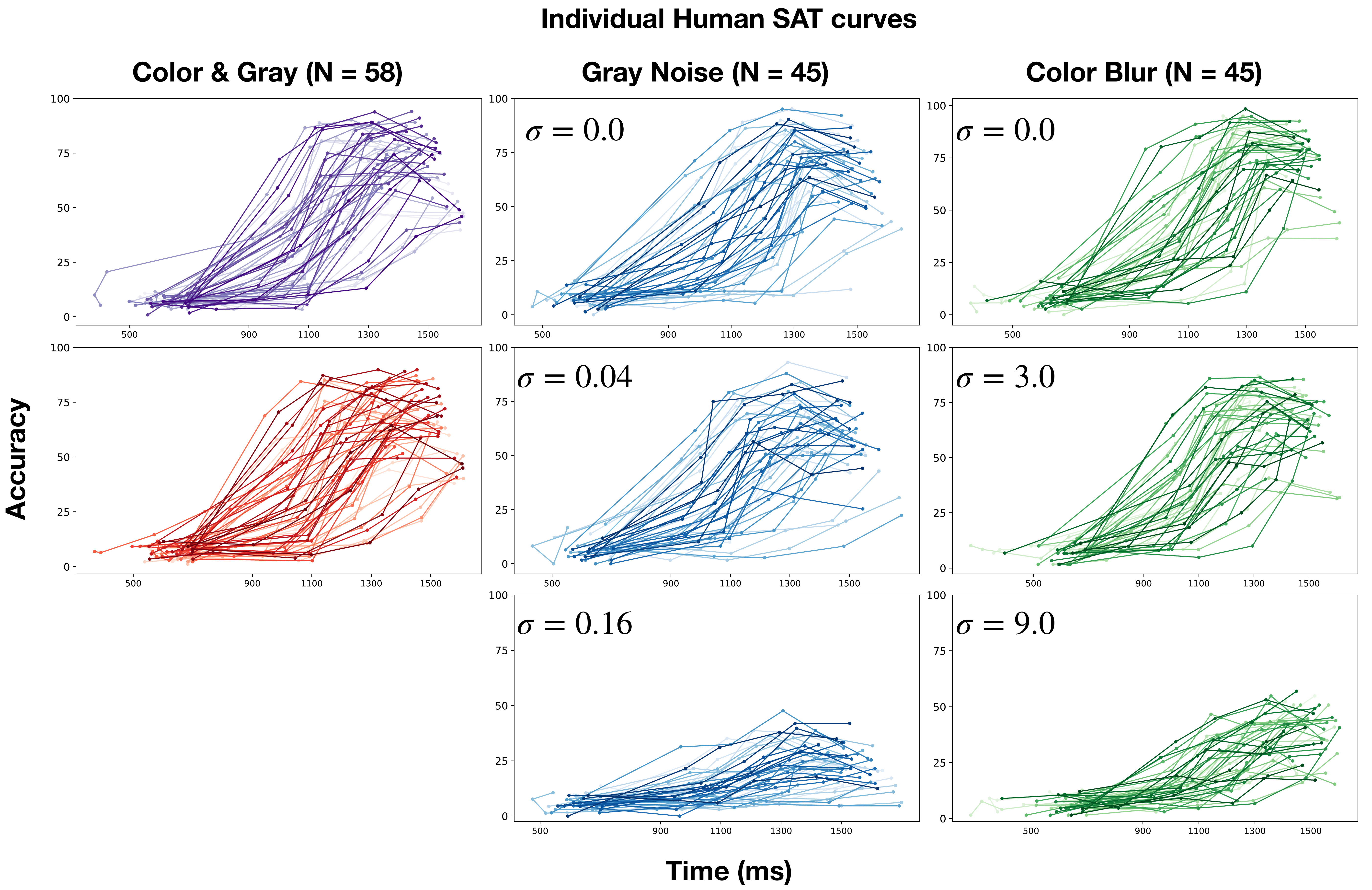}
   \caption{\emph{SAT curves for each human participant across all experimental conditions.} The curves are monotonic, showing the familiar sigmoidal increase of accuracy with time.}
   \label{fig:raw-human-sat}
\end{figure}

Figure \ref{fig:screenshots} illustrates the sequence of screens in our psychophysics experiment along with sample screenshots for each screen. Raw data of human accuracies vs reaction time, resulting from our experiments is shown in Figure \ref{fig:raw-human-sat}. Our dataset was collecting via Amazon Mechanical Turk \cite{crowston2012amazon} wherein each participant was paid \$20 for approximately an hour. 193 observers participated in our experiments. Thus, the total cost of data collection was \$3860. Data from 45 participants were discarded because more than 50\% of their responses were outside the required reaction time window of $\pm100$ ms from the beep. Therefore, we present all our results on the remaining 148 participants.

\section{Contrast Adjustment for Noise Experiments}
\begin{figure}[h]
   \centering
  \includegraphics[width=\linewidth]{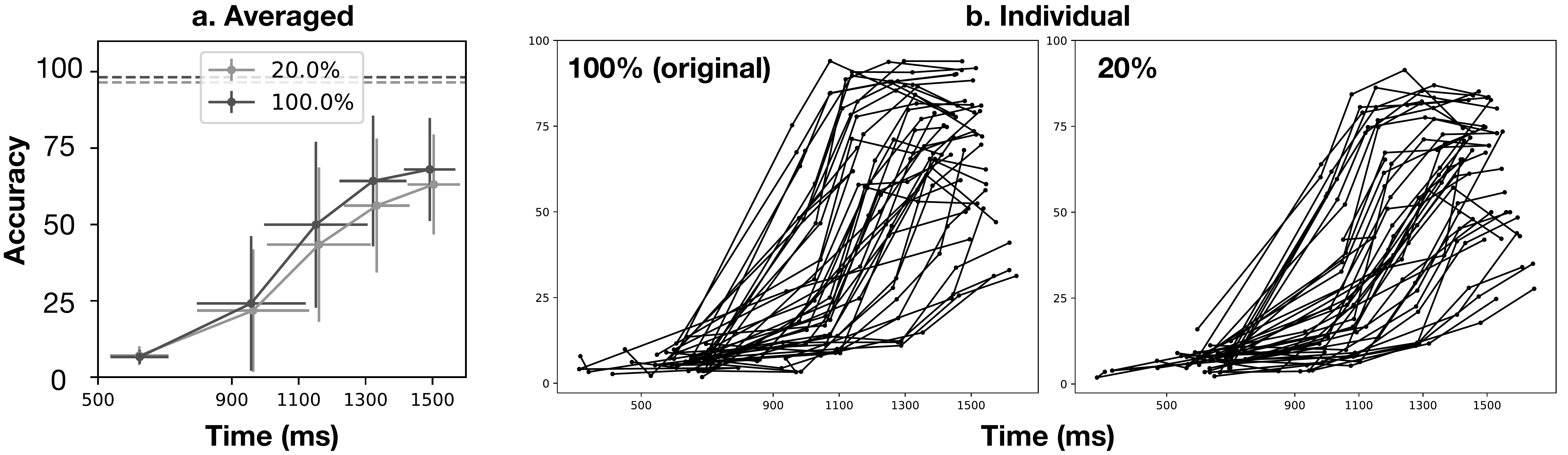}
   \caption{\emph{SAT curves when original (100\%) and low contrast (20\% of original) grayscale images were presented.} \textbf{a.} Averaged data for original and low-contrast conditions. Dotted lines denote untimed condition. \textbf{b.} Participant-wise SAT curves for original and low-contrast conditions. We notice a large variance in SAT curves but even the participants that required more time reached near-100\% in the untimed condition.}
   \label{fig:contrast}
\end{figure}

In our experiments evaluating humans and networks on noisy images, we perturbed images with Gaussian noise of zero mean and various standard deviation values. Since noise is additive and pixel values can only lie in a finite range (0.0-1.0), large perturbations to images could result in clipping which in-effect changes the noise distribution to salt-pepper. Thus, to make sure that our noise is Gaussian, we lower the contrast of the image to 20\% of the original before adding noise. Before doing so however, we have to verify that lower contrast does not deteriorate human performance, we ran an additional experiment testing people on original and 20\% contrast images. Results are shown in Figure \ref{fig:contrast}. We see no significant effect of lowering contrast on accuracy.

\section{Training and Evaluation Details}

For all networks, we used data augmentation during training based on standard techniques mentioned in \cite{huang2017multiscale}: images are horizontally flipped with probability 0.5, normalization based on channel means and standard deviation is also done.

\subsection{Cascaded Neural Network (CNet)}
For both serial and parallel CNet architectures, cross-entropy loss augmented with temporal difference (TD) learning is computed on a modified target $y_t$ at each timestep $t$.
$$y_t = (1-\lambda)\Big[\sum_{i=1}^{T-t} \lambda^{i-1}\hat{y}_{t+i}\Big] + \lambda^{T-t}y_{true}$$
where $\lambda \in [0,1]$ is a hyperparameter that governs weighting between the true target and future timestep predictions \cite{iuzzolino2021improving}. The model is trained over 120 epochs with batch size 128 and uses a Stochastic Gradient Descent (SGD) optimizer with Nesterov momentum 0.9 and weight decay 0.0005. We apply a multi-step decaying learning rate scheduled every 30 epochs with a starting rate of 0.01 and decay factor 0.2.

\subsection{Convolutional Recurrent Neural Network (ConvRNN)}
To prevent overfitting, the model was initialized with pre-trained ImageNet~\cite{russakovsky2015imagenet} weights and all layers before fully connected layers were frozen for subsequent training. The network was trained to optimize cross-entropy loss over classification targets using Adam optimizer with learning rate 0.005 and epsilon parameter 0.1. L2 regularization was applied throughout training with coefficient of $10^{-6}$. The model was trained for 100 epochs with a batch size of 128.

\subsection{Multi-Scale Dense Network (MSDNet)}
During training, MSDNet uses a cumulative cross-entropy classification loss computed over all early exits. The model is trained for 300 epochs and uses a Stochastic Gradient Descent (SGD) optimizer with a learning rate of 0.1 and batch size of 64. During evaluation, 5 approximately equally spaced exits are used.

\subsection{Scalable Neural Network (SCAN)}
During training, a loss function that combines a cross-entropy term (for classification) and a self-distillation term is computed and summed over all exits. The self-distillation helps improve accuracy by encouraging a low KL-divergence between the exit outputs and final output distributions, and is controlled using a self-distillation coefficient. In our experiments, SGD with a fixed learning rate of 0.1 and momentum factor of 0.9 is used to optimize network parameters. We use the default self-distillation coefficient of 0.5 and with a batch size of 128, for 50 epochs. During evaluation, 5 approximately equally spaced exits are used.

\section{Evaluation Metrics}

\subsection{Category-wise SAT Curves}
\begin{figure}[h]
   \centering
  \vspace{-0.4cm}
  \includegraphics[width=\linewidth]{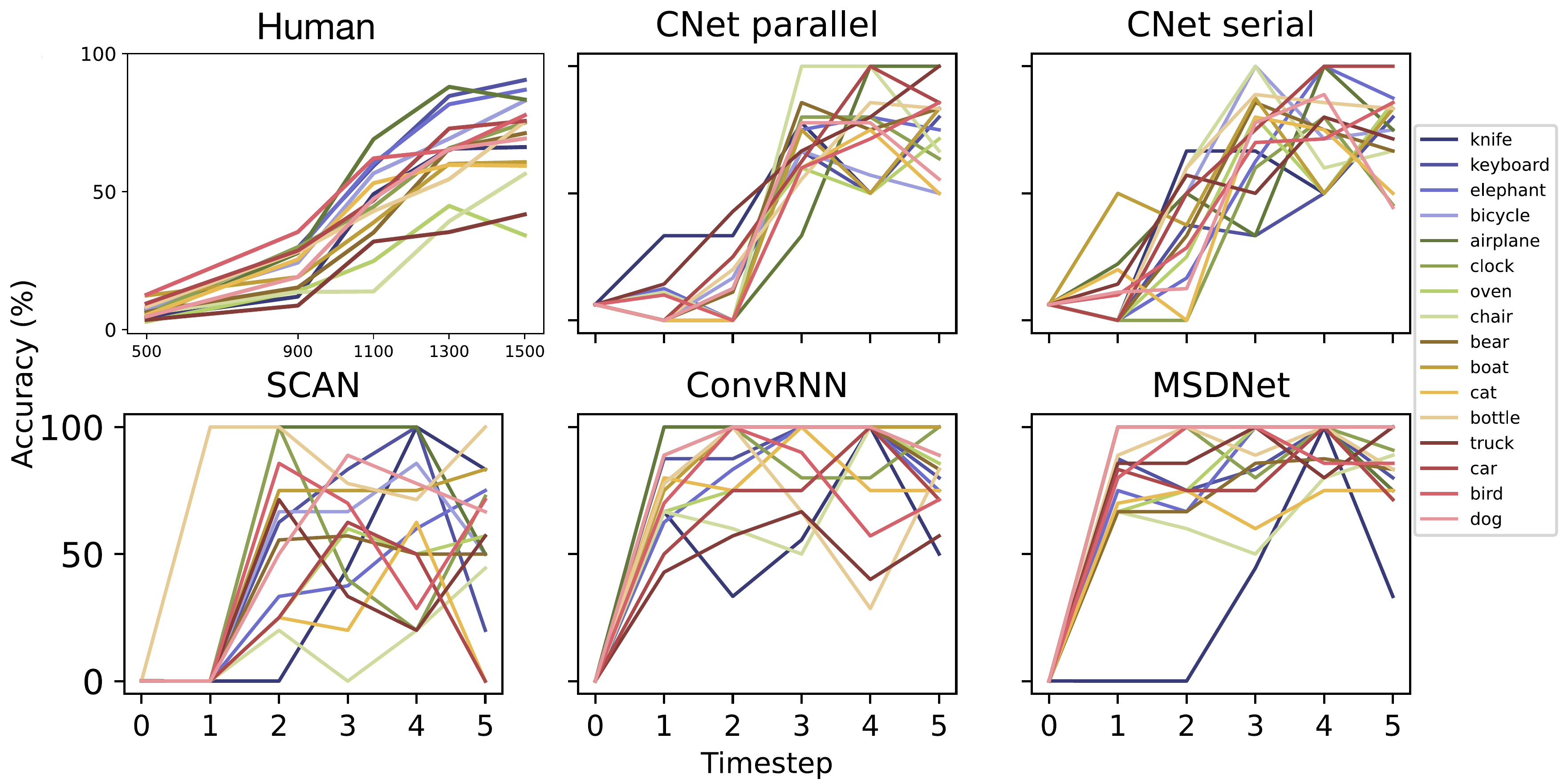}
   \caption{\emph{Category-wise accuracy-time curves for average human and all networks.} CNets, like humans, peak later and show similar behavior across all categories.}
   \label{fig:raw-category-correlation}
\end{figure}

Figure \ref{fig:raw-category-correlation} shows category-wise accuracy-time curves for the average human and all networks. In humans, the difference between category accuracies emerges only for large reaction time values. This is contrary to our expectation that easier categories would peak earlier than more difficult ones. Qualitatively, CNet-parallel captures the general trend of the human curves the best and CNet-serial follows closely behind. ConvRNN and MSDNet curves peak very early on and them along with SCAN show much larger differences between categories at early timesteps than humans do.\\

\subsection{Curve Steepness}
We proposed the steepness metric as a way to compare humans and networks on their ability to gracefully fail as time is decreased. We use mean curvature as a measure of steepness. A cumulative Weibull function $w(x)$ is first fit to the accuracy-time data points.
\begin{align*}
    w(t) = 1 - e^{-(t/\lambda)^k}
\end{align*}
where t is the input to the function (reaction time, in our case) and $\lambda$, $k$ are parameters to be fit. We use non-linear least squares to find the function that best fits each curve and then sample 20 equally spaced points for each. Finally, curvature $\kappa$ is found as shown below.

Speed is computed as the slope of the tangent to the curve at each sampled point. It is also the magnitude of the velocity vector (gradient of curve). This value will be used later on.
\[ \text{Speed} =\frac{d s}{d t}=|\mathbf{v}(t)|=\sqrt{\left(x^{\prime}\right)^{2}+\left(y^{\prime}\right)^{2}} \]

The tangent vector $\mathbf{T}$ to the curve is found by dividing the velocity vector $\mathbf{v}$ by speed. 
\[ \mathbf{v}=\frac{d s}{d t} \mathbf{T}, \quad \mathbf{T}=\frac{\mathbf{v}}{d s / d t} \]

Then, acceleration $\mathbf{a}(t)$ or the rate of change of the tangent's slope, is the derivative of velocity and can be expressed as:
\[ \mathbf{a}(t)=\frac{d^{2} s}{d t^{2}} \mathbf{T}+\kappa\left(\frac{d s}{d t}\right)^{2} \]

using which the curvature $\kappa$ can be obtained for plane curves as:
\[ \kappa=\frac{\left|x^{\prime \prime} y^{\prime}-x^{\prime} y^{\prime \prime}\right|}{\left(\left(x^{\prime}\right)^{2}+\left(y^{\prime}\right)^{2}\right)^{3 / 2}} \]

The above equations are implemented in code as follows.

\begin{lstlisting}
import numpy as np

def find_curvature(points):
	# calculate velocity
	x_t = np.gradient(points[:,0])
	y_t = np.gradient(points[:,1])
	vel = np.array([ [x_t[i], y_t[i]] for i in range(x_t.size)])

	# compute speed
	speed = np.sqrt(x_t * x_t + y_t * y_t)

	# compute tangent
	tangent = np.array([1/speed] * 2).transpose() * vel

	# find curvature
	ss_t = np.gradient(speed)
	xx_t = np.gradient(x_t)
	yy_t = np.gradient(y_t)
	curvature_val = np.abs(
	xx_t * y_t - x_t * yy_t
	) / (x_t * x_t + y_t * y_t)**1.5

	return np.mean(curvature_val), np.std(
	curvature_val
	)/np.sqrt(len(curvature_val))
\end{lstlisting}

\section{Compute Resources}
In order to train and test models for all our experiments, we used resources from an internal cluster at New York University. All networks were trained using 1-2 NVIDIA Tesla V100 GPUs requiring less than $100$ GB of memory. Training time for all networks was under 2 days. For each run of inference, we used either a single NVIDIA GeForce GTX 1080 Ti or Tesla V100 GPU.

\section{Datasheet for Human Data}
Data was collected for a timed object recognition task at varying levels of difficulty by having a rigid time regime and perturbations. Forced-choice responses of a test subject were recorded for different levels of perturbation in a timed setting.

\subsection{Motivation}
The core of everyday tasks like reading and driving is active object recognition. Attempts to model such tasks are currently stymied by the inability to incorporate time. People show a flexible tradeoff between speed and accuracy and this tradeoff is a crucial human skill. Deep neural networks have emerged as promising candidates for predicting peak human object recognition performance and neural activity. However, modeling the temporal dimension i.e., the speed-accuracy tradeoff (SAT), is essential for them to serve as useful computational models for how humans recognize objects. To this end, we here present the first large-scale (148 observers, 4 neural networks, 8 tasks) dataset of the speed-accuracy tradeoff (SAT) in recognizing ImageNet images.

The dataset was created at the Pelli Lab, Department of Psychology, New York University, via Lab.js for survey creation, JATOS for hosting and Amazon Mturk for crowdsourced data collection.

\subsection{Composition}
Dataset is divided into 3 sets based on what perturbation was used while collecting them. Table~\ref{tab:summary} provides a summary of dataset statistics. 

\begin{table}[h!]
\resizebox{0.75\textwidth}{!}{
\begin{tabular}{|l|r|r|r|l|}
\hline
\textbf{Perturbations} & \multicolumn{1}{l|}{\textbf{Participants}} & \multicolumn{1}{l|}{\textbf{Avg. Compl. (min)}} & \multicolumn{1}{l|}{\textbf{\#Trials}}                                                  \\ \hline
Color/Gray         & 58                                & 43.24                                  & 1100 \\ \hline
Noise          & 45                                 & 41.32                                  & 1100  \\ \hline
Blur         & 45                                 & 39.29                                  & 1100 \\ \hline
\end{tabular}
}
\caption{\emph{Summary of data collected via MTurk.}}
\label{tab:summary}
\end{table}

\subsubsection{Demographic Information}
We collect performance statistics from 148 observers (88 male, 59 female, 1 non-binary) whose ages ranged from 26 to 70 years, and who agreed to participate in an hour-long session. Each observer had normal or corrected-to-normal vision.

\subsubsection{Description of Raw Data}
Information collected via surveys is provided as JSON text files consisting of different fields important for measuring a speed-accuracy trade-off in an observer. The format of the data collection method can be found, in more detail, at lab.js: \texttt{\url{https://lab.js.org/}}. Important fields utilize to plot the speed-accuracy trade-off for an observer are:
\begin{itemize}
\item \textbf{url}: The first row of this column contains the `srid`, a unique ID for each participant in the experiment.

\item \textbf{order}: +1 indicates RT blocks were presented in ascending order and -1 means they were presented in descending order.

\item \textbf{sender}: Name of the page presented to the observer.

\item \textbf{filename}: Filename of image presented. 

\item \textbf{mode}: Value of perturbation applied to the image. 'c' for color, 'g' for grayscale and noise/blur value for noise and blur experiments.

\item \textbf{correctResponse}: Ground-truth category name of presented image.

\item \textbf{response}: Observer's category response.

\item \textbf{duration}: Reaction time of observer response.

\item \textbf{correct}: 1.0 if observer response was correct, else 0.0.

\end{itemize}

To understand how to process the raw data collected via surveys, please look at: {\texttt{\url{https://github.com/ajaysub110/satbench/tree/main/human_data_analysis}}}.

\subsubsection{Dependency}
This dataset of human observers was possible because of the public availability of the ImageNet dataset~\cite{russakovsky2015imagenet} and 16-class ImageNet subset \cite{geirhos2018generalisation}. There are no restrictions on using it for research purposes. Examples of perturbations on ImageNet images can be found in the main manuscript. 
For the ImageNet and 16-class ImageNet licenses, please visit: \texttt{\url{https://www.image-net.org/index.php}} and \texttt{\url{https://github.com/rgeirhos/generalisation-humans-DNNs}}.

\begin{figure}[h]
      \centering
    \includegraphics[width=0.83\linewidth]{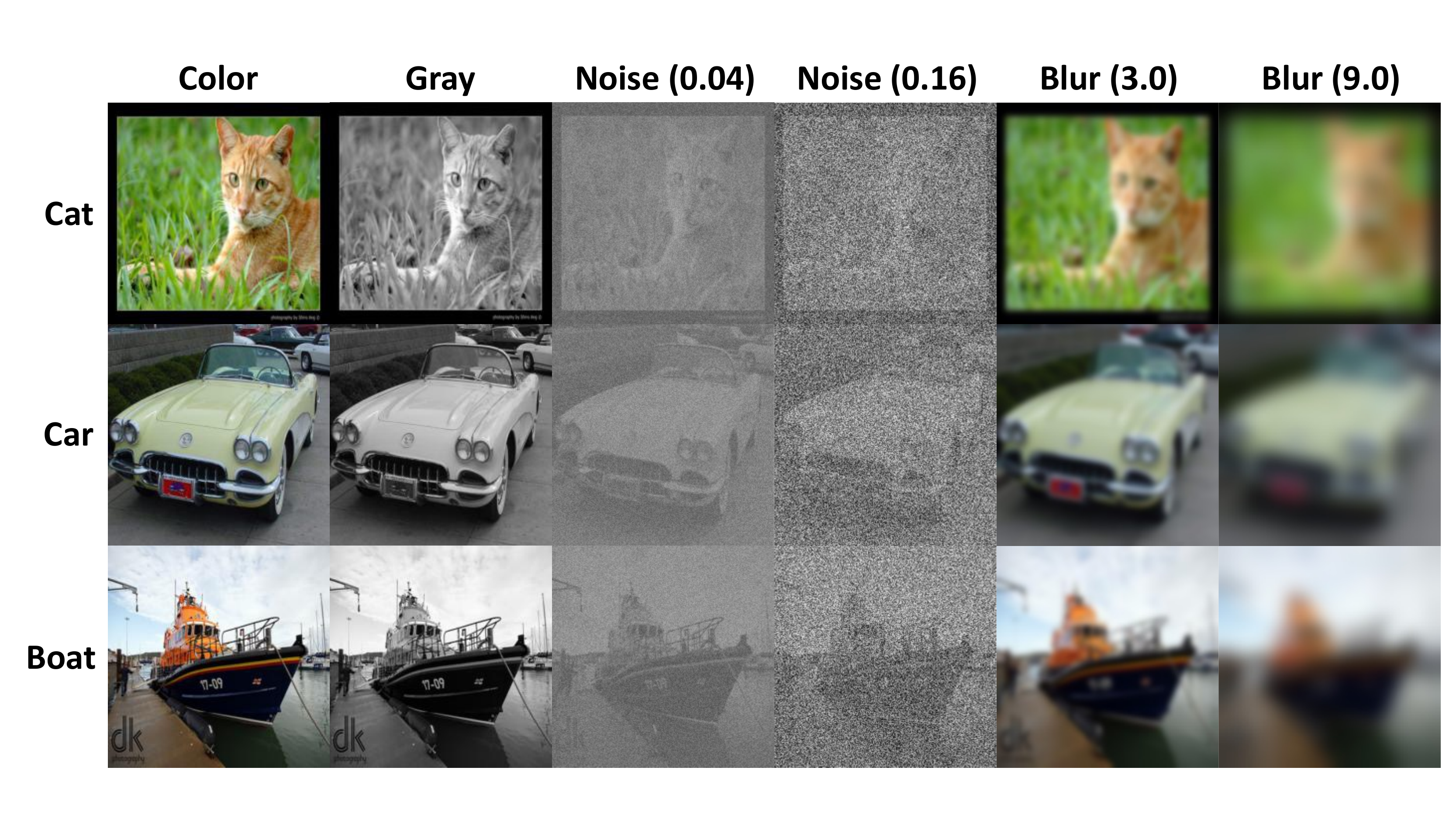}
     \caption{\emph{Example images from the ImageNet dataset \cite{russakovsky2015imagenet} along with visualizations of image perturbations considered for human subject experiments -- grayscale conversion, image blurring and noise.} Numbers in parentheses correspond to standard deviations for 0-mean Gaussian distributions. Replicated from main paper.}
      \label{fig:samples}
\end{figure}

\subsubsection{Participants}
We collect age and gender from participants taking the survey. No other information is collected. This dataset cannot be used to calculate any sub-populations, or identify individuals directly or indirectly. 

\subsection{Collection Process}
To collect data, we used lab.js~\cite{henninger_felix_2020_3953072} to design our surveys, JATOS~\cite{Lange_2015} to host them, and MTurk to pay participants 20\$ per hours for their efforts, with a total of \$3860 with all fees.

\subsubsection{Survey Design}
Prior to presenting the stimuli, a sample of 1,100 images was taken randomly from the ImageNet validation dataset and different perturbations were added to create a sample set which was added to the survey. 

The stimuli were presented via JATOS survey via worker links to each observer. A standard IRB approved (IRB-FY2016-404) consent form was signed before collecting the data by each observer, and demographic information (age and gender) was collected. For different perturbations, observers were given specific instructions to complete the survey.

Prior to the study, subjects were instructed to click on the buttons corresponding to each of the 16 object categories: airplane, bear, bicycle, bird, boat, bottle, car, cat, chair, clock, dog, elephant, keyboard, knife, oven and truck. They also had a training run where they were asked to categorize 50 images.

Stimuli images were scaled to 400x400 pixels for optimal viewing. The survey was designed on five fixed viewing conditions (blocks) of 500 ms, 900 ms, 1100 ms, 1300 ms and 1500 ms with a tolerance of 100 ms each. Outside of these tolerance values, trials were discarded.

For color, noise and blur surveys, each time condition block consisted of 210 trials plus training (1100 trials in total). At the end of the time-limit for a trial, a beep sounded within 60 ms of which the observer had to enter their category decision via virtual button-click after which feedback was given: if they were quick, slow or perfect while clicking the button.

\subsection{Time for Collection}
Designing of a survey took around 1 month. Using MTurk for getting data was faster and data for 33 observers was done within 1.5 months. An IRB-approved form was signed before the start of each survey and a participant had the right to withdraw from the survey at any time. 

\subsubsection{Hosting the Survey}
We used JATOS to host and deploy surveys created using lab.js. Hosted surveys can be accessed at:
\begin{itemize}
\item Noise: \url{http://64.225.11.86/publix/171/start?batchId=174&generalMultiple}
\item Blur: \url{http://64.225.11.86/publix/170/start?batchId=173&generalMultiple}
\item Color/Gray: \url{http://64.225.11.86/publix/167/start?batchId=170&generalMultiple}
\end{itemize}

\subsection{Preprocessing}

 Dataset was collected in the form of surveys and has information related to reaction time and noise. The Jupyter notebooks provided showcase how to process the dataset and create a benchmark for modeling human reaction time. Each psychometric function or data collected from a single observer is in the form of a JSON text file which can be imported as a dataframe using pandas~\cite{reback2020pandas,mckinney-proc-scipy-2010} in python language.
 You can checkout how to process data here: 
 \texttt{\url{https://github.com/ajaysub110/satbench/tree/main/human_data_analysis}}.

\subsection{Uses}
The main purpose of this dataset is to provide a benchmark for models exhibiting anytime prediction ability or the ability to effectively trade-off speed and accuracy. Our work compares neural networks with humans on the speed-accuracy trade-off (SAT) task of object recognition and is a fundamental step to understanding various public health issues, such as dyslexia. Possible applications also include object detection in resource-constrained devices and self-driving cars.

\subsubsection{Visualizations}
Visualizations of observer SAT curves across all conditions are shown in Figure \ref{fig:raw-human-sat}. Further analysis can be done by using notebooks provided in \texttt{\url{https://github.com/ajaysub110/satbench/tree/main/human_data_analysis}}.

\section{Additional Dataset Information}
\subsection{Accessing Our Dataset} 
Our dataset is publicly available at \texttt{\url{https://osf.io/2cpmb/}} and along with visualization notebooks and a detailed description of its contents at \texttt{\url{https://github.com/ajaysub110/satbench}}. We guarantee that all results and observations from the paper can be replicated using the code and data available in the repository. 

\subsection{Author Statement} 
We confirm that we will abide by the rules of the Creative Commons (CC) License and will take responsibility for any violation of rights.

\subsection{Hosting, licensing, and maintenance plan} 
Our dataset is hosted on OSF where it is available for free under the Creative Commons (CC) License. The authors will continue to provide any necessary maintenance.

\end{document}